# AI-DECISION SUPPORT SYSTEM INTERFACE USING CANCER RELATED DATA FOR LUNG CANCER PROGNOSIS

Asim Leblebici[1], Omer Gesoglu[2], Yasemin Basbinar[3]

1. Dokuz Eylul University, Institute of Health Sciences, Department of Translational Oncology, Izmir, Turkey

2. Uskudar American Academy, Istanbul; Dokuz Eylul University, Institute of Oncology, Department of Translational Oncology, Pre-graduated Research Group, Izmir, Turkey.

3. Dokuz Eylul University, Institute of Oncology, Department of Translational Oncology, Izmir, Turkey

**ABSTRACT**

Until the beginning of 2021, lung cancer is known to be the most common cancer in the world. The disease is common due to factors such as occupational exposure, smoking and environmental pollution. The early diagnosis and treatment of the disease is of great importance as well as the prevention of the causes that cause the disease. The study was planned to create a web interface that works with machine learning algorithms to predict prognosis using lung cancer clinical and gene expression in the GDC data portal.

**INTRODUCTION**

Lung cancer, of which smoking is one of the leading factors, ranks first in cancer-related deaths. However, when it is detected at an initial stage, the chance of treatment of the disease increases. Lung cancer starts when cells from structurally normal lung tissue proliferate out of need and control, forming a mass (tumor) in the lung. The mass formed here primarily grows in its environment, and in more advanced stages, it causes damage by spreading to surrounding tissues or to distant organs (liver, bone, etc.) through circulation. It accounts for 12-16 percent of all cancers and 17-28 percent of cancer-related deaths. Moreover, it ranks first in cancer-related deaths in both women and men.

**MATERIAL-METHOD**

In the study, known as TCGA, with its new name GDC data portal was used. GDC data portal is a big data environment that contains different types of data on 33 cancer types [1].

*Data collection tool*

Tcgabiolinks package provides access to GDC data portal data in R environment, provides functions related to analysis and visualization. Clinical and gene expression data were downloaded from the colon adenocarcinoma-TCGA-COAD project using the Tcgabiolinks package [2].

*Analysis of clinical data*

Clinical data were evaluated by Kaplan-Meier survival analysis.

*Analysis of Gene Expression data*

In gene expression data, primary solid tumor subtype was selected and samples containing more than one data were eliminated. Gene expression analysis was performed to show gene expression changes between groups based on right and left column information. In this analysis, fold change was calculated and student t-test analysis was applied to calculate p-values. The p-values were corrected using the "FDR-False positive rate" method. For each pair of groups, statistically significant gene lists were created by taking genes with a folding ratio of > 1.0 and FDR <0.05.

*Gene Enrichment Analysis*

The pathways that were statistically significant in the KEGG[3] and Cancer Hallmark [4] databases were selected in the gene enrichment analysis results with the EnrichR package [5]. PathfindR package was used for gene-pathway visualization [6].

*Machine Learning Algorithms*

Genes in the selected pathways were evaluated using machine algorithm methods. Confusion matrix table were created from test sets created by 10-fold cross validation method.

*Decision support tool - Shiny Web Interface*

The clinical and transcriptomic data obtained were transformed into a decision support web interface using the R-Shiny package [7].

**RESULTS**

*Clinical parameters*

TCGA-LUAD clinical data were analyzed by Kaplan-Meier survival analysis. In the stage information, survival shows a statistically significant decrease as the stage progresses. Higher survival is also seen in the initial stages in T, N, M staging types.

Table 1: TCGA-LUAD clinical parameters with Kaplan-Meier survival analysis results

| Parameters | | Total N | N of Events | Median Est. | Std. Error | 95% LCL | 95% UCL | p |
|---|---|---|---|---|---|---|---|---|
| Stage | Stage i | 270 | 65 | 87,33 | 25,101 | 38,133 | 136,527 | <0,001 |
| | Stage ii | 119 | 54 | 40,3 | 7,007 | 26,566 | 54,034 | |
| | Stage iii | 81 | 46 | 26,9 | 4,720 | 17,649 | 36,151 | |
| | Stage iv | 26 | 16 | 27,53 | 7,091 | 13,631 | 41,429 | |
| ajcc_pathologic_t | T1 | 168 | 43 | 77,27 | 17,628 | 42,719 | 111,821 | 0,003 |
| | Other | 333 | 138 | 42,93 | 3,415 | 36,236 | 49,624 | |
| ajcc_pathologic_n | N0 | 325 | 86 | 77,27 | 14,519 | 48,812 | 105,728 | <0,001 |
| | Other | 167 | 93 | 31,73 | 3,012 | 25,827 | 37,633 | |
| ajcc_pathologic_m | M0 | 335 | 129 | 50,03 | 3,944 | 42,299 | 57,761 | 0,005 |
| | M1 | 25 | 15 | 32,53 | 7,393 | 18,040 | 47,020 | |
| Dimension | <0.7 | 136 | 42 | 77,27 | 18,351 | 41,302 | 113,238 | 0,050 |
| | >=0.7 | 249 | 95 | 41,17 | 4,688 | 31,981 | 50,359 | |
| Morphology | 8140/3 | 303 | 129 | 40,3 | 2,954 | 34,509 | 46,091 | 0,001 |
| | Others | 201 | 54 | 89,37 | 25,659 | 39,078 | 139,662 | |
| Malignancy | No | 424 | 154 | 50,03 | 4,873 | 40,478 | 59,582 | 0,085 |
| | Yes | 80 | 29 | 35,77 | 7,311 | 21,440 | 50,100 | |
| Primary diagnosis | Adeno | 303 | 129 | 40,3 | 2,954 | 34,509 | 46,091 | 0,001 |
| | Other | 201 | 54 | 89,37 | 25,659 | 39,078 | 139,662 | |
| Cigarette per day | <2.2 | 169 | 54 | 57,5 | 9,619 | 38,646 | 76,354 | 0,833 |
| | >=2.2 | 176 | 66 | 48,47 | 6,801 | 35,141 | 61,799 | |
| Years smoked | <32 | 90 | 23 | 59,27 | 9,933 | 39,801 | 78,739 | 0,035 |
| | >=32 | 99 | 41 | 33,17 | 4,465 | 24,419 | 41,921 | |

*Differential expression gene set (DEGs) enrichment results*

Gene expressions of the groups were compared using the survival information on the TCGA-LUAD database. Significant genes were queried in KEGG pathway and Cancer Hallmark databases. In the Kegg pathway results, it was found that the expression of "CACNA1A", "GABRA2" genes decreased and "GRIA2", "GRIA1" genes increased in "Nicotine addiction" term. In the Hallmark pathway results, it was found that the expression of "COL2A1", "SLC12A32", "EPHA5" genes decreased and "TENM2", "SERPINA10", "KRT13", "KCNQ2", "CDH16", "KRT5", "WNT16", "SCGB1A1" genes increased in "Kras signaling" term.

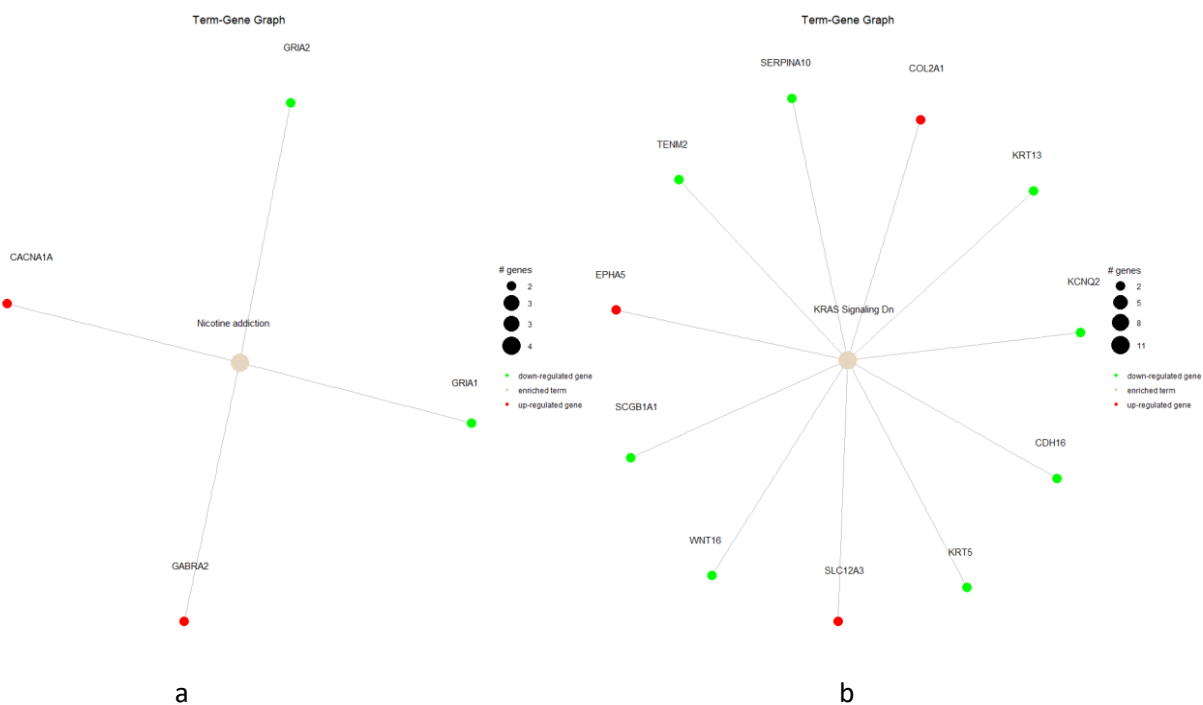

a    b

Figure 1a: Up/down differential expression gene set in KEGG pathway results
Figure 1b: Up/down differential expression gene set in Cancer Hallmark pathway results

*Machine learning algorithm results*

Machine learning algorithm results were created with 4 different sets of attributes. The results are shown in table 2. The feature set formed by clinical data with "Nicotine addiction" parameters was the cluster with the highest accuracy rate 70.2% with decision tree algorithm. The highest sensitivity result (88.7%) was obtained with the decision tree algorithm in three different feature sets.

Table 2: Various selected parameters scenario for cancer risk classification

| Parameters | Algorithms | Sensitivity | Specificity | AUC | Accuracy |
|---|---|---|---|---|---|
| Clinical Parameters | Decision Tree | 87,7% | 63,6% | 61,0% | 69,0% |
| | Random Forest | 84,4% | 58,8% | 66,3% | 68,6% |
| | Naive Bayes | 78,8% | 47,6% | 69,2% | 69,2% |
| | SVM | 77,3% | 59,4% | 59,0% | 63,9% |
| Clinical & Nicotine addiction genes | Decision Tree | 88,7% | 62,0% | 59,6% | 70,2% |
| | Random Forest | 82,8% | 56,1% | 66,3% | 68,6% |
| | Naive Bayes | 77,3% | 45,5% | 69,6% | 69,0% |
| | SVM | 78,2% | 61,5% | 58,4% | 63,7% |
| Clinical & KRAS signaling genes | Decision Tree | 88,7% | 65,8% | 60,4% | 68,8% |
| | Random Forest | 81,3% | 59,9% | 65,2% | 66,3% |
| | Naive Bayes | 77,6% | 48,1% | 69,8% | 68,2% |
| | SVM | 79,8% | 59,4% | 60,2% | 65,5% |
| All parameters | Decision Tree | 88,7% | 65,2% | 59,1% | 69,0% |
| | Random Forest | 78,5% | 56,7% | 64,3% | 65,7% |
| | Naive Bayes | 77,9% | 53,5% | 69,0% | 66,5% |
| | SVM | 81,0% | 58,3% | 61,3% | 66,7% |

*Shiny Web UI*

Using the obtained results, R Shiny package, https://gesogluomer.shinyapps.io/luad/ link was created. The interface of the web page is shown in figures 2 and 3. The visual interface is user-friendly with a click-to-run form. Shiny interface has an infrastructure that provides decision support by selecting the clinical characteristics and expression of genes. Includes decision result, algorithm detail, complexity table and figure on the screen.

Figure 2: Web interface running machine learning algorithms in the background using expression and clinical data

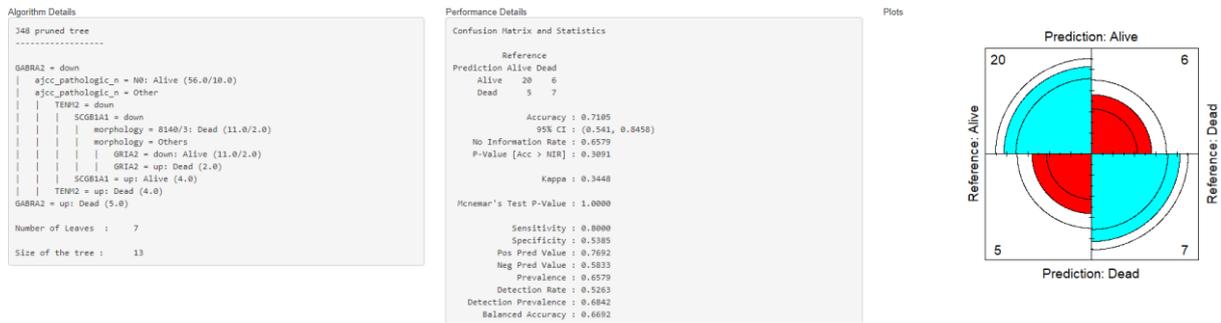

Figure 3: Algorithm details of the web interface

**DISCUSSION**

As a result of the study, the targeted web interface was created. Results were obtained using both statistical methods and bioinformatics analysis. The lack of some clinical data and the availability of accessibility constitute the limitation of the study. In order to increase the performance of algorithm results, it is planned to continue with targets such as increasing data, different approaches in feature selection, filling the missing data with appropriate methods, algorithm optimization.

# REFERENCES

[1]   F. S. Collins, "The Cancer Genome Atlas ( TCGA )," *Online*. 2007, doi: 10.1038/nature07943.

[2]   A. Colaprico *et al.*, "TCGAbiolinks: An R/Bioconductor package for integrative analysis of TCGA data," *Nucleic Acids Res.*, 2016, doi: 10.1093/nar/gkv1507.

[3]   M. Kanehisa, Y. Sato, M. Kawashima, M. Furumichi, and M. Tanabe, "KEGG as a reference resource for gene and protein annotation," *Nucleic Acids Res.*, 2016, doi: 10.1093/nar/gkv1070.

[4]   A. Liberzon, C. Birger, H. Thorvaldsdóttir, M. Ghandi, J. P. Mesirov, and P. Tamayo, "The Molecular Signatures Database Hallmark Gene Set Collection," *Cell Syst.*, 2015, doi: 10.1016/j.cels.2015.12.004.

[5]   M. V. Kuleshov *et al.*, "Enrichr: a comprehensive gene set enrichment analysis web server 2016 update," *Nucleic Acids Res.*, 2016, doi: 10.1093/nar/gkw377.

[6]   E. Ulgen, O. Ozisik, and O. U. Sezerman, "PathfindR: An R package for comprehensive identification of enriched pathways in omics data through active subnetworks," *Front. Genet.*, 2019, doi: 10.3389/fgene.2019.00858.

[7]   W. Chang, J. Cheng, J. Allaire, Y. Xie, and J. McPherson, "Package ' shiny ': Web Application Framework for R," 2020.